\documentclass[10pt, conference]{IEEEtran}

\usepackage[latin1]{inputenc}  
\usepackage{graphicx,url}
\usepackage{multirow}  
\usepackage{hhline}
\usepackage{graphicx}
\usepackage{subfigure,amssymb,amsmath}
\usepackage{color,soul}
\usepackage[dvipsnames]{xcolor}
\usepackage{bbm}
\usepackage{float}

\makeatletter  
\newif\if@restonecol  
\makeatother

\usepackage[linesnumbered,ruled,vlined]{algorithm2e}
\usepackage{algpseudocode}  
\usepackage{amsmath}  
\usepackage{fancyhdr} 
\usepackage{tabularx}
\usepackage{comment} 

\usepackage{balance}

\renewcommand\thetable{\arabic{table}}
\IEEEoverridecommandlockouts

\usepackage{etoolbox}
\makeatletter
\patchcmd{\@makecaption}
  {\scshape}
  {}
  {}
  {}
\makeatother

\begin{document}


\title{Dynamically meeting performance objectives for multiple services on a service mesh}

\author{\IEEEauthorblockN{Forough Shahab Samani \IEEEauthorrefmark{2} and 
 Rolf Stadler\IEEEauthorrefmark{2}}

 \IEEEauthorblockA{\IEEEauthorrefmark{2} Dept. of Computer Science, KTH Royal Institute of Technology, Sweden}
 {Email: \{foro, stadler\}@kth.se 
}
\\\today
}

\maketitle


\thispagestyle{plain}
\pagestyle{plain}

\begin{abstract}
We present a framework that lets a service provider achieve end-to-end management objectives under varying load. Dynamic control actions are performed by a reinforcement learning (RL) agent. Our work includes experimentation and evaluation on a laboratory testbed where we have implemented basic information services on a service mesh supported by the Istio and Kubernetes platforms. We investigate different management objectives that include end-to-end delay bounds on service requests, throughput objectives, and service differentiation. These objectives are mapped onto reward functions that an RL agent learns to optimize, by executing control actions, namely, request routing and request blocking. We compute the control policies not on the testbed, but in a simulator, which speeds up the learning process by orders of magnitude. In our approach, the system model is learned on the testbed; it is then used to instantiate the simulator, which produces near-optimal control policies for various management objectives. The learned policies are then evaluated on the testbed using unseen load patterns.  
\end{abstract}


\section{Introduction}
\label{sec:introduction}

End-to-end performance objectives for a service are difficult to achieve on a shared and virtualized infrastructure. This is because the service load often changes in an operational environment, and service platforms do not offer strict resource isolation, so that the resource consumption of various tasks running on a platform influences the service quality. 

In order to continuously meet performance objectives for a service, such as bounds on delays or throughput for service requests, the management system must dynamically perform control actions that re-allocate the resources of the infrastructure. Such control actions can be taken on the physical, virtualization, or service layer, and they include horizontal and vertical scaling of compute resources, function placement, as well as request routing and request dropping.

The service abstraction we consider in this paper is a directed graph, where the nodes represent processing functions and the links communication channels. This general abstraction covers a variety of services and applications, such as a network slice on a network substrate, a service chain on a softwarized network, a micro-service based application, or a pipeline of machine-learning tasks. We choose the service-mesh abstraction in this work and apply it to micro-service based applications.

In this paper, we propose a framework for achieving end-to-end management objectives for multiple services that concurrently execute on a service mesh. We apply reinforcement learning (RL) techniques to train an agent that periodically performs control actions to reallocate resources. A management objective in this framework is expressed through the reward function in the RL setup. We develop and evaluate the framework using a laboratory testbed where we run information services on a service mesh, supported by the Istio and Kubernetes platforms \cite{Istio},\cite{K8}. We investigate different management objectives that include end-to-end delay bounds on service requests, throughput objectives, and service differentiation. 

Training an RL agent in an operational environment (or on a testbed in our case) is generally not feasible due to the long training time, which can extend to weeks, unless the state and action spaces of the agent are very limited. We address this issue by computing the control policies in a simulator rather than on the testbed, which speeds up the learning process by orders of magnitude for the scenarios we study. In our approach, the RL system model is learned from testbed measurements; it is then used to instantiate the simulator, which produces near-optimal control policies for various management objectives, possibly in parallel. The learned policies are then evaluated on the testbed using unseen load patterns (i.e. patterns the agent has not been trained on). 

We make two contributions with this paper. First, we present an RL-based framework that computes near-optimal control policies for end-to-end performance objectives on a service graph. This framework simultaneously supports several services with different performance objectives and several types of control operations. \\
Second, as part of this framework, we introduce a simulator component that efficiently produces the policies. Through experimentation, we study the tradeoff of using the simulator versus learning the policies on a testbed. We find that while we lose some control effectiveness due to the inaccuracy of the system model we gain by significantly shortening the training time, which makes the approach suitable in practice. To the best of our knowledge, this paper is the first to advocate a simulation phase a part of implementing a dynamic performance management solution on a real system. 

Note that, when developing and presenting our framework, we aim at simplicity, clarity, and rigorous treatment, which helps us focus on the main ideas. For this reason, we choose a small service mesh for our scenarios (which still includes key complexities of larger ones), we consider only two types of control actions, etc. Our plan is to refine and extend the framework in future work, as we lay out in the last section of the paper.

\section{Problem formulation and approach}
\label{sec:problem_formulation_and_approach}


\begin{table}[!htbp]
\centering
\color{black}
\caption{Notation for formalizing the problem.}
\label{tab:notation}
\begin{tabular}{|l|c|}
\hline
\textbf{Concept} & \textbf{Notation} \\\hline \hline
Service index & $i$ \\\hline
Node index & $j$ \\\hline
Offered load of service $i$ & $l_i$ \\\hline
Carried load of service $i$ & $l_i^c=l_i(1-b_i)$\\\hline
Utility of service $i$ & $u_i$ \\\hline
Response time objective & $O_i$ \\\hline
Response time & $d_i$\\\hline
Routing weight of service $i$  & $p_{ij}$\\
towards node $j$&\\\hline
Blocking rate & $b_i$\\\hline

\hline
\end{tabular}
\end{table} 

\begin{figure}[!htbp] 
  \centering
  \subfigure[Directed graph of a service mesh]{\label{fig:service_mesh}
     \includegraphics[scale=0.55]{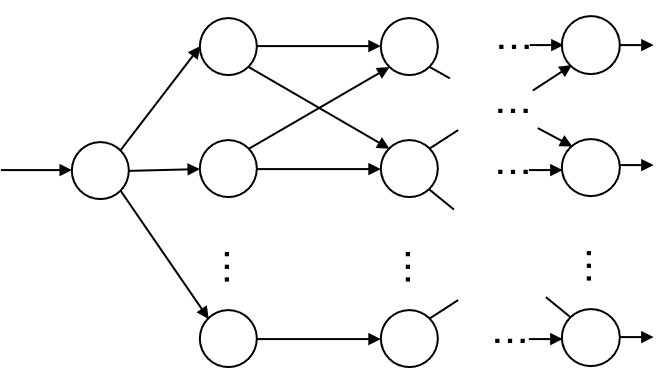}}\hspace{0.15cm}
  \subfigure[Use case in this work]{\label{fig:service_mesh_sub}
     \includegraphics[scale=0.6]{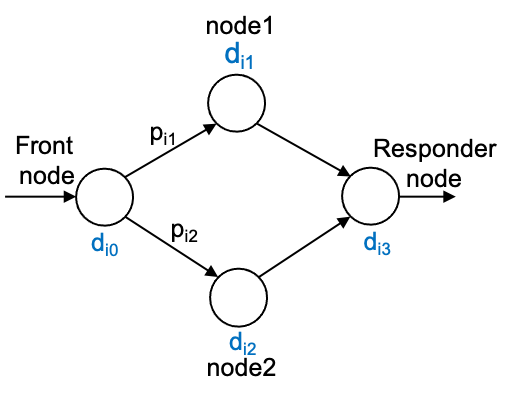}} 

  \caption{(a) General service mesh; each processing node runs one or more microservices; links are communication channels for service requests. (b) Service mesh of the use case; $d_{ij}$ is the processing delay of a request of service $i$ on node $j$; $p_{ij}$ is the routing weight of request $i$ towards node $j$.}
  \label{fig:graphs}
\end{figure}

We consider application services built from microservice components, whereby each component performs a unique function. A service request traverses a path on a directed graph whose nodes offer microservices. Figure \ref{fig:service_mesh} shows a directed graph with a general topology of a microservice architecture, which we call a service mesh. A service is realized as a contiguous subgraph on the service mesh. 

While the framework we present in this paper applies to the general service mesh of Figure \ref{fig:service_mesh}, we focus the discussion and experimentation on the smaller configuration shown in Figure \ref{fig:service_mesh_sub}. An incoming service request from a client is processed first by the front node, before being routed to one of the two processing nodes. The processing result is sent to the responder node which sends it to the client.

We consider two services  $S_i$ with different resource requirements on the processing nodes. In the configuration in Figure \ref{fig:service_mesh_sub}, both processing nodes can process requests from both services. We express the service quality in terms of end-to-end delay $d_i$ of a service request, the carried load of a service $l_i^c$ (which we also call the throughput of the service), or the utility $u_i$ generated by a service.

In order to control the service quality, our framework includes control actions. In this work, we consider request routing expressed by $p_{ij} \in [0,1]$ which indicates the fraction of requests of service $i$ routed to processing node $j$. In addition, we consider blocking service requests at the front node at the fraction $b_i \in [0,1]$. 

Central to this work are \emph{management objectives}, which capture the end-to-end performance objectives for the services on a given service mesh. These objectives include client requirements, as well as provider priorities. To present and evaluate our framework, we focus on the following three management objectives. See Table 1 for notation.

\textbf{Management Objective 1 (MO1)}: the response time of a request of service $i$ is upper bounded by $O_i$ and the overall carried load is maximized: 

\begin{equation} \label{eq:mo1}
\text{maximize   } \sum_i l_i^c \text{ while } 
d_i<O_i
\end{equation}

\textbf{Management Objective 2 (MO2)}: the response time of a request of service $i$ is upper bounded by $O_i$ and the sum of service utilities is maximized:

\begin{equation} \label{eq:mo2}
\text{maximize   } \sum_i u_i \text{ while } d_i<O_i
\end{equation}

\textbf{Management Objective 3 (MO3)}: the response time of a request of service $i$ is upper bounded by $O_i$, the carried load $l_i^c$ of service $i$ is maximized, while service $k$ is prevented from starving (i.e., by specifying a lower threshold $l_{min}$ for service $k$):

\begin{equation} \label{eq:mo3}
\begin{split}
\text{maximize   } l_i^c \text{ while }d_i<O_i \text{ and } l_k^c>l_{min}  \  i\neq k
\end{split}
\end{equation}

For the management objective M1, M2, or M3, we solve the following problem for the configuration in Figure \ref{fig:service_mesh_sub}. Given the offered load of service $i$, and the response time $d_i$, we need to find the control parameters $p_{ij}$ and $b_i$. We obtain these parameters through reinforcement learning where they represent the optimal policy. 

Following the reinforcement learning approach, we formalize the above problem as a Markov Decision Process (MDP), which is a well-understood model for sequential decision making \cite{puterman2014markov}\cite{sutton2018reinforcement}. The elements of this formalization for the scenario in Figure \ref{fig:service_mesh_sub} are:


\noindent \textbf{State space:} The state at time $t$ includes the response time and offered load of the services running on the service mesh: $s_t \in \{(d_{i,t},l_{i,t}) \in \mathbb{R}^{2m}| \ i= 1,...,m\} = \mathcal{S}$, where m is the number of services.  

\noindent \textbf{Action/control space:} The action at time $t$ is a vector of the control parameters: $a_t \in \{b_{i,t}, p_{ij,t} | \ i=1,...,m, j=1,2\} = \mathcal{A}$. 

\noindent \textbf{System model}: This model provides the new state when specific action is taken in a given state: 
\begin{equation}
\label{eq:system_model}
    (d_{i,t+1},l_{i,t+1}) = f(d_{i,t}, l_{i,t} , b_{i,t} , p_{ij,t} | \ j = 1,2) \ \  i=1,...,m
\end{equation}
Since we train the policy on a known load pattern, the system model we must learn can be simplified, as we argue in Section \ref{sec:scenarios_and_evaluation_set_up}.

\noindent \textbf{Reward function}: The reward function expresses the management objective in the reinforcement learning context. The reward functions of all three management objectives are detailed in Section \ref{sec:scenarios_and_evaluation_set_up}.

Learning an effective or close-to-optimal policy on a real system is often infeasible due to the time that it takes for the system to reach a new stable state after an action. On our testbed, training a reinforcement learning agent can require days or even weeks as we show in Section \ref{sec:evaluation}. For this reason, our framework includes two systems. The first is a lab testbed, on which the system model is learned. The second is a simulation system, which is instantiated with the system model and on which the optimal policy is learned. The learned policy is then transferred to the testbed where it can be evaluated \cite{hammar_stadler_simulation_emulation}.  

\section{Testbed and request generation}
\label{sec:testbed_description}



Our testbed at KTH includes a server cluster connected through a Gigabit Ethernet switch. The cluster contains nine Dell PowerEdge R715 2U servers, each with 64 GB RAM, two 12-core AMD Opteron processors, a 500 GB hard disk, and four 1 Gb network interfaces. The tenth machine is a Dell PowerEdge R630 2U with 256 GB RAM, two 12-core Intel Xeon E5-2680 processors, two 1.2 TB hard disks, and twelve 1 Gb network interfaces. 
All machines run Ubuntu Server 18.04.6 64 bits and their clocks are synchronized through NTP \cite{NTP}. The orchestration layer and the service mesh are realized using Kubernetes (K8) \cite{K8} and Istio \cite{Istio}.

\begin{figure}[!htbp]
 \centering
 \includegraphics[scale=0.45]{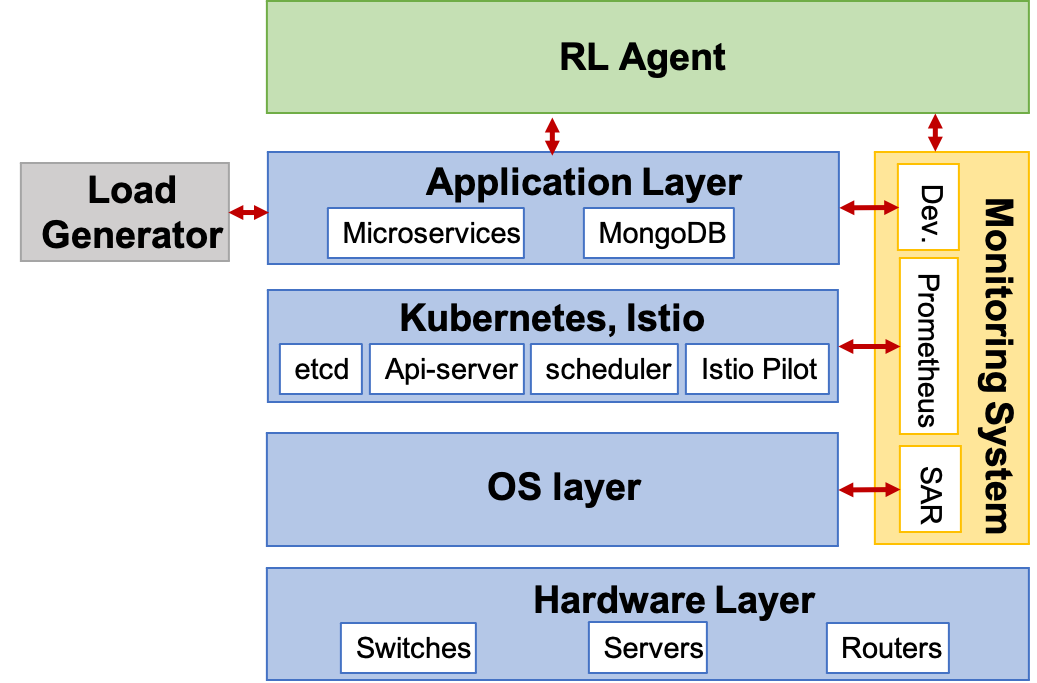}
 \caption{Software stack of the testbed at KTH}
 \label{fig:testbed}
\end{figure}

On top of the Istio service mesh, we implemented two information services, which we call service $1$ and service $2$. Both services, upon receiving request with a key, return a corresponding data item from a database. The difference between the services is the structure of data items and the size of the databases. 

\begin{figure}[!htbp]
 \centering
 \includegraphics[scale=0.4]{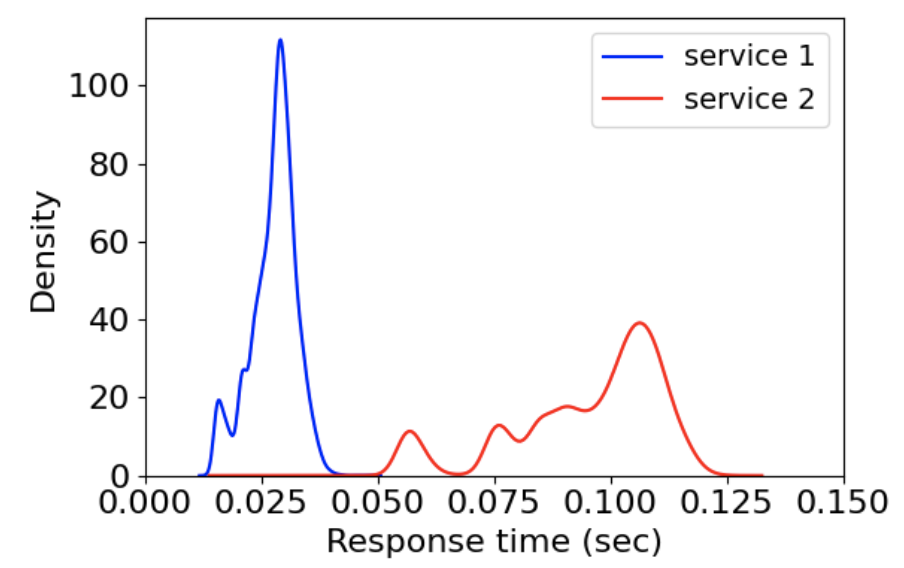}
 \caption{Response time distribution of service $1$ and service $2$.}
 \label{fig:services_response_dist}
\end{figure}

Figure \ref{fig:services_response_dist} shows the distribution of the response times of service requests for both services in a specific scenario. 

\begin{figure}[!htbp]
 \centering
 \includegraphics[scale=0.40]{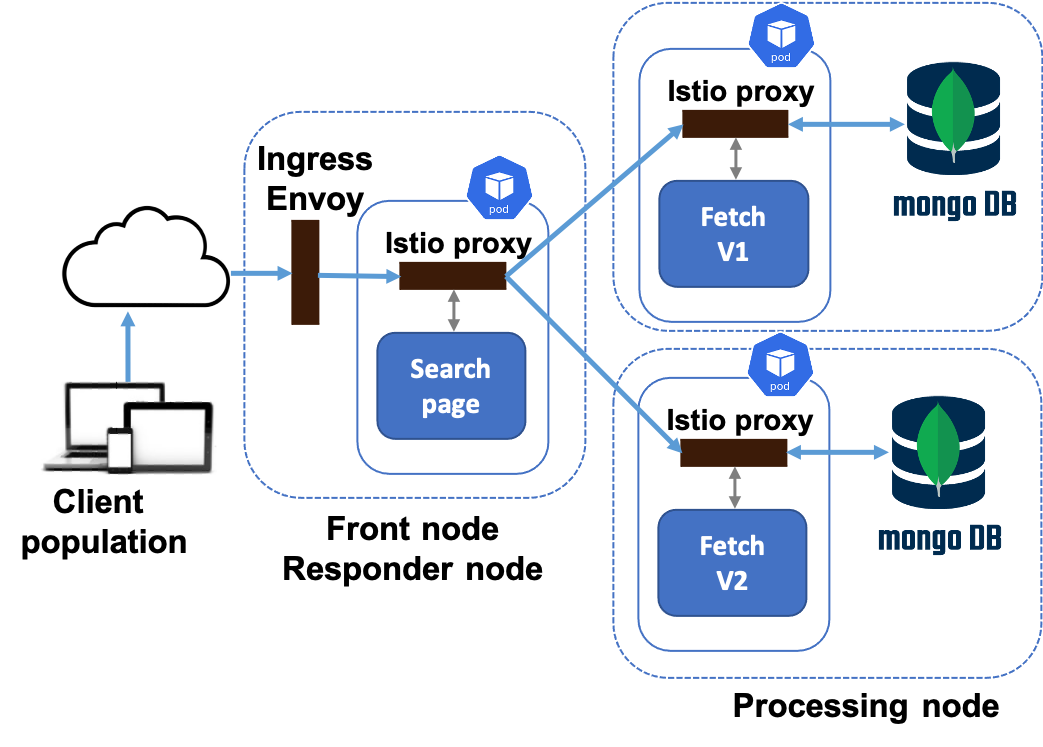}
 \caption{The architecture of the microservice-based application deployed over the service mesh. See Figure \ref{fig:service_mesh_sub}.}
 \label{fig:application}
\end{figure}

Figure \ref{fig:application} shows the implementation of the services on the testbed following the configuration given in Figure 1(b). The front node and the responder node in Figure 1(b) are realized as a single node in Figure \ref{fig:application}. All nodes are implemented in Python \cite{python} using Flask \cite{flask}. The front node provides the web user interface and the other two nodes provide the content for the information service. Each processing node is implemented as a Kubernetes pod \cite{pods}.


We implemented a load generator in order to emulate a client population. It is driven by a stochastic process that creates a stream of service requests. We realize two load patterns with this generator.\\ 
The \textit{random load pattern} produces a stream of requests at the rate of $l_i(t)\sim\mathbbm{U}_{\{5,10,15,20\}}$ requests/second. It changes at every time step to a value drawn uniformly at random from ${\{5,10,15,20\}}$. A time step is $5$ seconds on the testbed. \\
The \textit{sinusoidal load pattern} produces a stream of requests at the rate of $l_i(t) = 12.5 + 7.5 \times \sin (\frac{2\pi}{T} t + \phi)$ requests/second.  


\section{Scenarios and evaluation set up}
\label{sec:scenarios_and_evaluation_set_up}

We study three scenarios that help us evaluate our approach to efficiently learn effective policies on the testbed for different management objectives. At the core of each scenario is a management objective that captures the intention of the service provider and a reward function that reflects the management objective in the reinforcement learning framework. 

\begin{figure}[!htbp] 
  \centering
  \subfigure[$r_1(s_t,a_t)$]{\label{fig:sc1_reward_2d_s1}
     \includegraphics[scale=0.25]{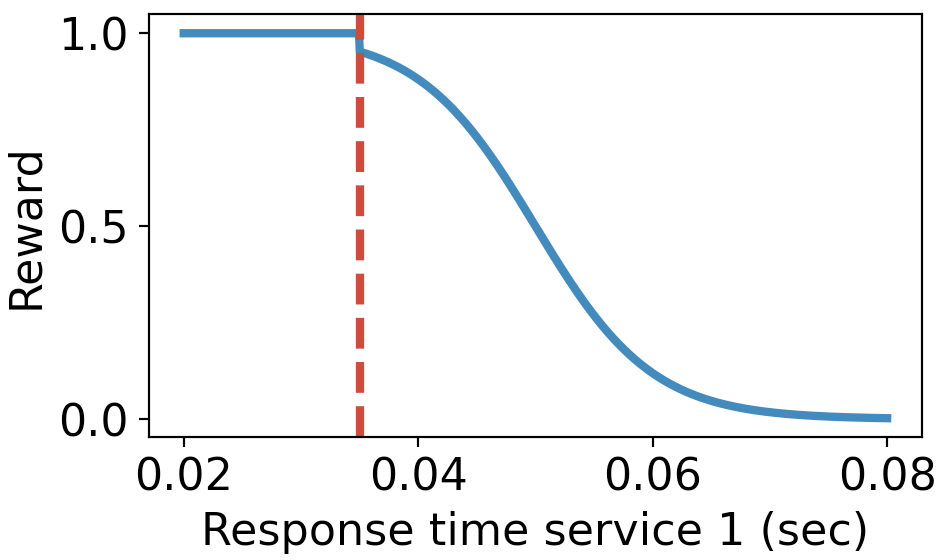}}
  \subfigure[$r_2(s_t,a_t)$]{\label{fig:sc1_reward_2d_s2}
     \includegraphics[scale=0.25]{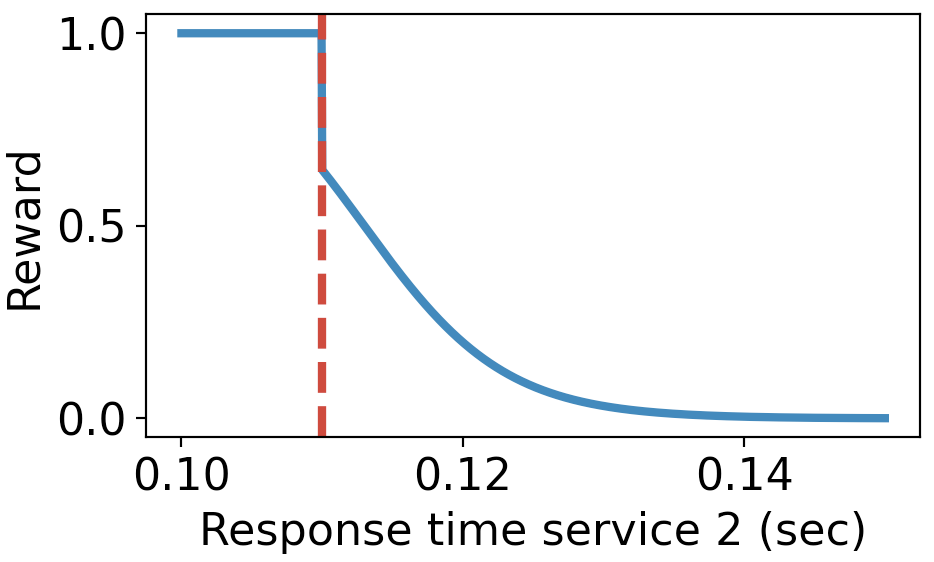}} 

  \caption{The components of the reward functions for MO1 and MO2.}
  \label{fig:sc1_rewards_2d}
\end{figure}

\textbf{Scenario 1:} We run the two services (i.e., $m=2$) under management objective 1, where the RL agent attempts to maximize the joint throughput while enforcing constraints on the response times under changing load. The management objective MO1 \\
$$\text{ maximize } (l_1^c + l_2^c) \text{ while } d_1<O_1 \text{ and } d_2<O_2$$ 

\noindent is implemented through the reward function \\
$$r(s_t,a_t)= l_1^c \times r_1(s_t,a_t) + l_2^c \times r_2(s_t,a_t)$$ 

\noindent whereby $r_1$ and $r_2$, depicted in Figure \ref{fig:sc1_rewards_2d} are functions based on $tanh$, capturing the reward of the average response time during time step $t$ for service $1$ and service $2$, respectively.



\textbf{Scenario 2:} We run the two services (i.e., $m=2$) under management objective 2, where the RL agent attempts to maximize the sum of the utility functions $u_1$ and $u_2$ while enforcing constraints on the response times under changing load. The management objective MO2 for this case is\\
$$\text{maximize } (u_1+u_2) \text{ while } d_1<O_1 \text{ and } d_2<O_2$$

In this scenario, the service provider gives priority to service $2$ over service $1$ by setting $u_2=5 \times l_2^c$ and $u_1= l_1^c$. 
This objective is achieved by maximizing the reward function \\
$$r(s_t,a_t)= l_1^c \times r_1(s_t,a_t) + 5 \times l_2^c \times r_2(s_t,a_t)$$

where $r_1$ and $r_2$ are defined above. 


\textbf{Scenario 3:} We run the two services (i.e., $m=2$) under management objective 3, where the RL agent attempts to maximize the throughput of service $2$ while enforcing constraints on the response time of service $2$ and guaranteeing a minimum throughput of $l^{min}_{1}$ to service $1$ under changing load. The management objective MO3:  \\
$$\text{maximize } l_2^c \text{ while } d_2<O_2 \text{ and }l_1^c>l^{min}_{1}$$

Like in scenario 2, the service provider gives priority to service $2$ over service $1$ but prevents service $1$ from starving. 
This objective is achieved by maximizing the reward function \\
$$r(s_t,a_t)= l_2^c \times\bigl(r_3(s_t,a_t) + r_2(s_t,a_t)\bigl)$$
\noindent where $r_3$ is a function based on $tanh$, capturing the reward of the carried  load during time step $t$ for each service. 


Common to all scenarios is the reinforcement learning model. The state of the system at time $t$ is $$s_t=(d_{1,t},d_{2,t},l_{1,t},l_{2,t})$$ 
The action the agent takes at time $t$ is $$a_t = (p_{11,t},p_{21,t},b_{1,t},b_{2,t})$$
We discretize the action space, which facilitates estimating the highest achievable performance of the system. We discretize every component of the action space by allowing six values so that $$\mathcal{A} = \{[0,0,0,0],[0,0,0,0.2],[0,0,0,0.4],...,[1.0,1.0,1.0,1.0]\}$$

\textbf{System model:} We learn the system model through testbed observation and supervised learning. Measurements on the testbed suggest that the delay at time step $t+1$ does not depend on the delay at time step $t$. Therefore, the system model in Equation \ref{eq:system_model} simplifies to\\
\begin{equation}
\label{eq:system_simple}
    (d_{1,t+1},d_{2,t+1}) = f(l_{1,t}, l_{2,t} , p_{11,t}, p_{21,t}, b_{1,t}, b_{2,t})
\end{equation}


\textbf{Optimal policy:} To obtain the optimal policy for a given state, we compute the rewards of all possible actions using the system model. We call the highest reward the \emph{optimal reward}, and the actions giving this reward determine the \emph{optimal policy}. 

\textbf{Evaluation metric:} The metric measures to which extent the management objective is met in a specific scenario. We define it as the average of the obtained reward divided by the optimal reward and call it ANR (Average Normalized Reward): 

\begin{equation}
\label{eq:metric}
    ANR = \frac{1}{T}\sum_{t=1}^T \frac{\text{obtained reward(t)}}{\text{optimal reward(t)}}
\end{equation}

The value of this metric is between $0$ and $1$. $1$ means the management objective is always met. Sometimes, we measure the performance of the agent for specific time step $t$ and call this metric Normalized Reward ($NR$) at time $t$.

\textbf{Learning the system model:} We run both services on the testbed under varying load, whereby at each time step $t$ the load for both services changes independently, taking values uniformly at random from the set $\{5,10,15,20\}$ requests per second. Each time step is $5$ seconds. In addition, actions are taken randomly from the action space $\mathcal{A}$. We run the system for $45\,343$ time steps and collect the data $(d_{1}(t),d_{2}(t), l_1(t), l_2(t) , p_{11}(t), p_{21}(t), b_{1}(t), b_{2}(t))$ as a trace. 

Taking this trace, we learn the system model using random forest regression \cite{BR01}\cite{RF_sklearn} with input $(l_1, l_2 , p_{11}, p_{21}, b_{1}, b_{2})$ and target $(d_{1},d_{2})$. The same system model is used to learn the control policies for all management objectives. 

\textbf{Learning the control policies:} We use the Stables Baselines3 library \cite{baselines3} to implement the RL agent, selecting the PPO method. To train the agent, we choose a neural network of size $[64, 64]$, a batch size of $64$, a distance between updates of $1\,024$ steps, the learning rate $\alpha = 0.001$, and the discount factor $\gamma = 0$. We find that the number of iterations for the agent to converge is scenario-specific and is in the order of $10\,000$. 

\section{Evaluation results and discussion}
\label{sec:evaluation}

\textbf{Training the RL agent:} Control policies are learned on a simulator which acts as the environment for the RL agent. At each time step, the simulator determines the current load and executes the function of the system model (Equation \ref{eq:system_simple}). Note that this simulator is very different from a discrete event simulator. \\
When the simulation starts, the agent is initialized with a random policy. During a simulation run, we track the learning process by monitoring the reward and the $ANR$ (Equation \ref{eq:metric}). We terminate the simulation when these metrics have sufficiently converged. We need to train the agent for each management objective separately, because a control policy is specific to an objective. However, control policies can be computed in parallel. 

\textbf{Evaluation of learned policies:} First, we evaluate a learned policy on the simulator. Given a management objective and a related scenario, we run the scenario for both load patterns (see Section \ref{sec:testbed_description}) and measure the $ANR$. Since the random load pattern has been used for training the agent, we expect close to optimal performance. The sinusoidal load pattern has not been seen by the agent and allows us to assess how the policy generalizes to unseen load. \\
Second, we evaluate the learned policy on the testbed. We run each scenario on the testbed for both load patterns. Like above we measure the performance of the RL agent by computing the $ANR$. Comparing the performance of the RL agent on the simulator with that on the testbed gives us an indication of the accuracy of the learned system model. 

\begin{figure}[!htbp]
 \centering
 \includegraphics[scale=0.35]{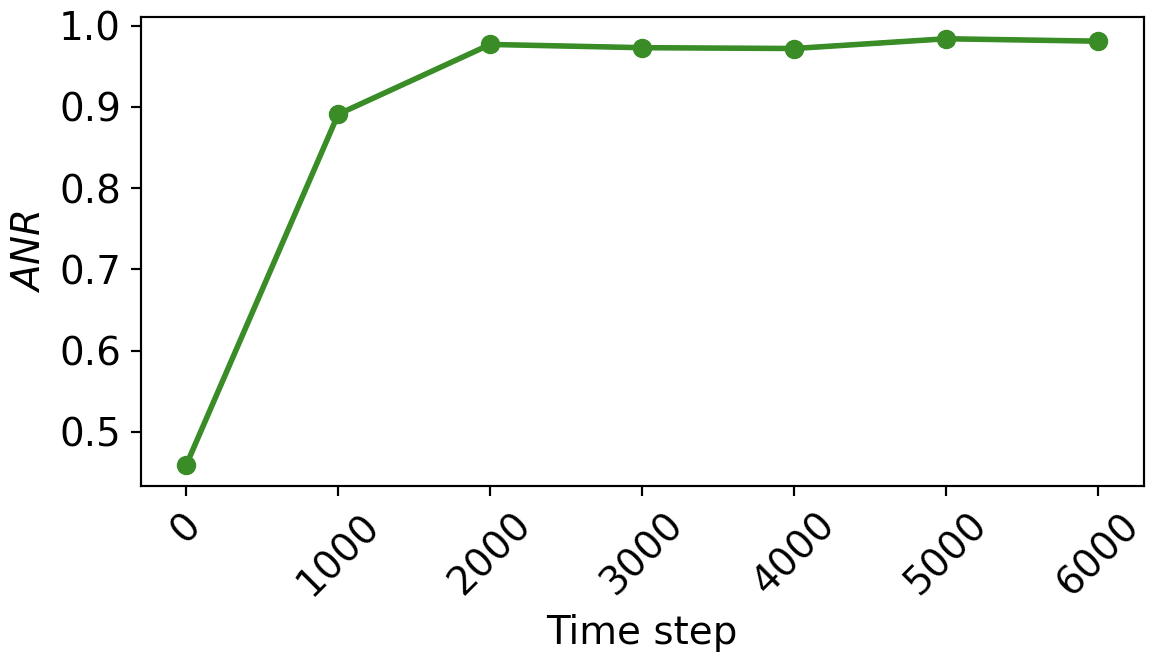}
 \caption{The learning curve of the RL agent for performance metric $ANR$ on the simulator running Scenario 1.}
 \label{fig:learning_curve}
\end{figure}

\textbf{Evaluation Scenario 1:}
We run the simulator for Scenario 1. For each time step, the simulator computes the current load for both services, executes the RL agent to obtain the control vector, determines the related reward, and computes the new state. During the simulation run, we compute the $ANR$ and produce the learning curve shown in Figure \ref{fig:learning_curve}. We see from the curve that the agent learns an increasingly effective policy. The $ANR$ value approaches $0.98$ after some $2\,000$ time steps.

\begin{figure}[!htbp] 
  \centering
  \subfigure[Offered and carried load]{\label{fig:sc1_sim_seen_ls}
     \includegraphics[scale=0.40]{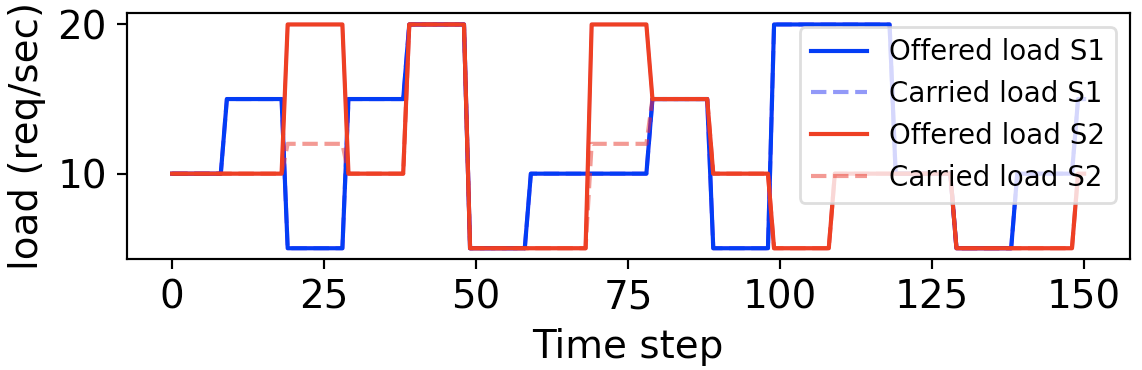}} \hspace{0.15cm}
  \subfigure[Performance of the RL agent]{\label{fig:sc1_sim_seen_nr}
     \includegraphics[scale=0.40]{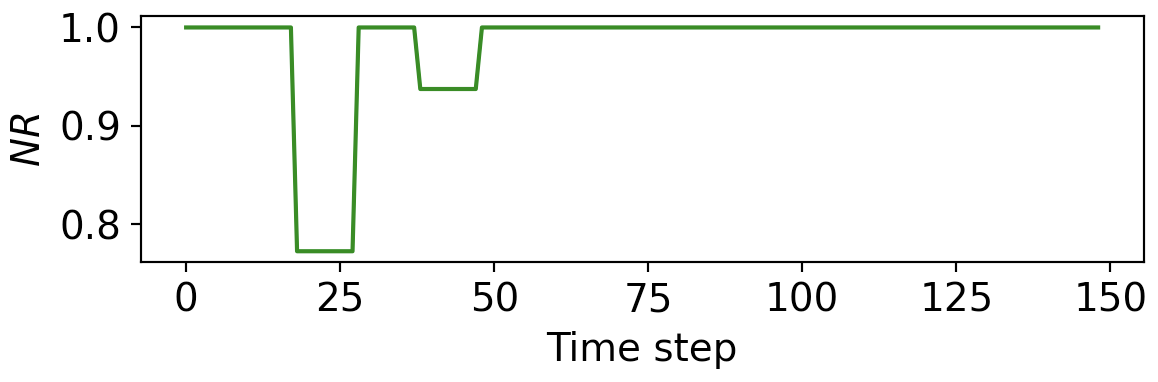}}
     
  \caption{(a) Offered and carried load of both services for random load in Scenario 1. (b) Performance of the RL agent in $NR$. Measurements taken from simulator.}
  \label{fig:sc1_sim_seen_load}
\end{figure}

Figure \ref{fig:sc1_sim_seen_ls} shows the offered and carried load for both services during a time interval of the $150$ time steps. We observe that the agent blocks some of the requests of service $2$ but none of service $1$. Figure \ref{fig:sc1_sim_seen_nr} shows the performance of the agent in $NR$ for the same time period. We see that there are time intervals where the policy is not optimal, i.e. $NR < 1$. We explain this by the fact that the learned policy did not fully converge during the training period and by the probabilistic behavior of the PPO agent.

\begin{figure}[!htbp] 
  \centering
  \subfigure[Offered and carried load]{\label{fig:sc1_sim_unseen_ls}
     \includegraphics[scale=0.4]{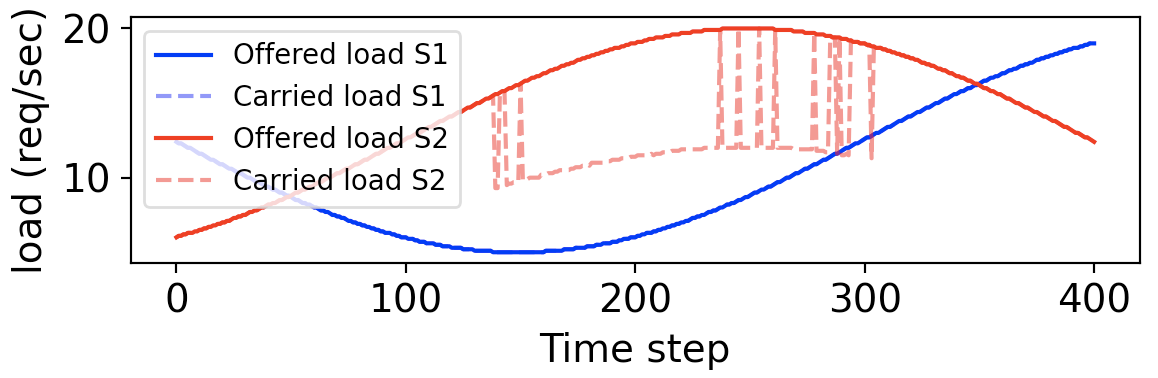}} \hspace{0.15cm}
  \subfigure[Performance of the RL agent]{\label{fig:sc1_sim_unseen_nr}
     \includegraphics[scale=0.4]{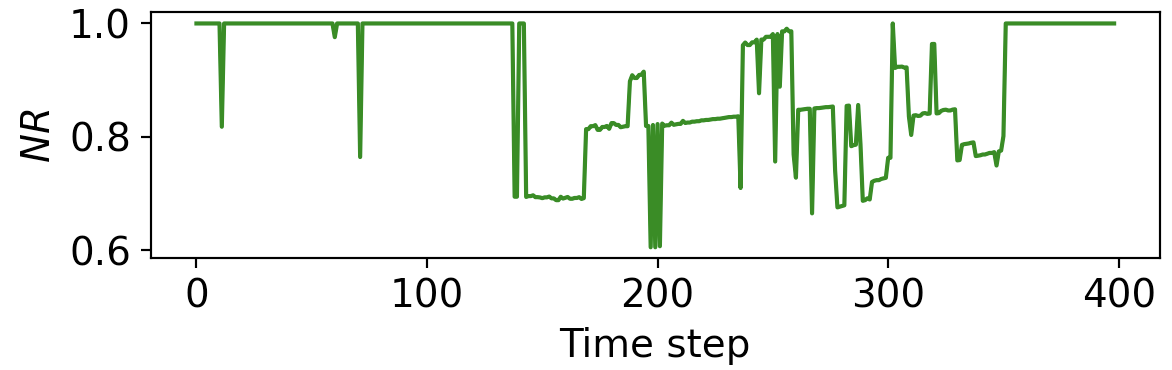}}
  \caption{(a) Offered and carried load of both services for sinusoidal load in Scenario 1. (b) Performance of the RL agent in $NR$. Measurements taken from simulator.}
  \label{fig:sc1_sim_unseen_load}
\end{figure}

Figure \ref{fig:sc1_sim_unseen_load} relates to Scenario 1 under the unseen, sinusoidal load pattern during an interval of $400$ time steps. This experiment allows us to evaluate to which extent the learned policy generalizes to an unseen load pattern. Similar to Figure \ref{fig:sc1_sim_seen_load}, Figure \ref{fig:sc1_sim_unseen_ls} shows the offered and carried load for both services, and Figure \ref{fig:sc1_sim_unseen_nr} depicts the performance of the agent for the same time period. We observe that, as with the random load pattern, the agent sometimes blocks requests of service $2$ when the system experiences high load. Also, we see that the $NR$ value can be lower than $1$ and can change significantly over time. We explain the sudden changes with the discretized action space, which limits the  actions the agent can take. 

\begin{figure}[!htbp] 
  \centering
  \subfigure[$NR$ for random load]{\label{fig:sc1_emu_same_nr}
     \includegraphics[scale=0.4]{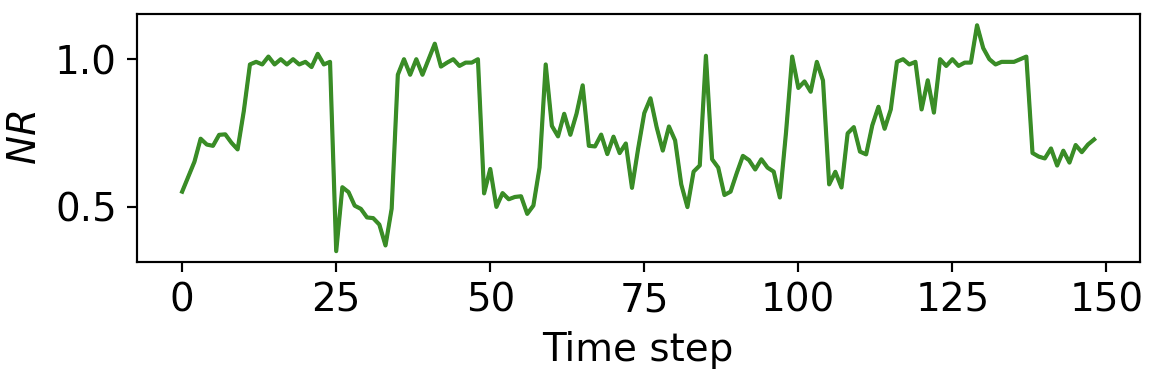}} \hspace{0.15cm}
  \subfigure[$NR$ for sinusoidal load]{\label{fig:sc1_emu_sin_nr}
     \includegraphics[scale=0.4]{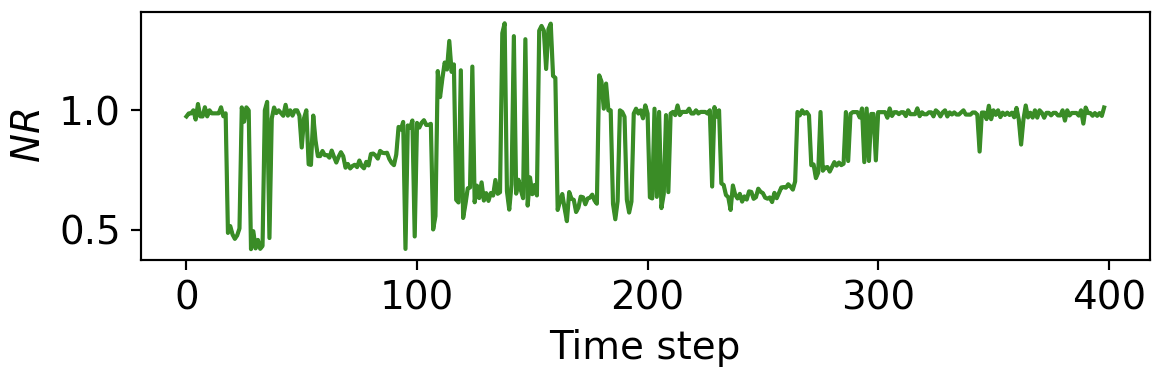}}
  \caption{Performance of the learned policy on the testbed for the both load patterns.}
  \label{fig:sc1_emu_nrs}
\end{figure}

Figure \ref{fig:sc1_emu_nrs} shows the performance of the agent on the testbed for Scenario 1. Figure \ref{fig:sc1_emu_same_nr} gives the $NR$ values for a time window under the random load pattern, and Figure \ref{fig:sc1_emu_sin_nr} shows the same metric under the sinusoidal load pattern. Surprisingly, the $NR$ is sometimes larger than 1, which seems to contradict its definition (Equation \ref{eq:metric}). When checking the data, we find that the optimal reward is wrongly computed if this happens, because of the inaccuracy of the system model. 

\begin{figure}[!htbp]
 \centering
 \includegraphics[scale=0.4]{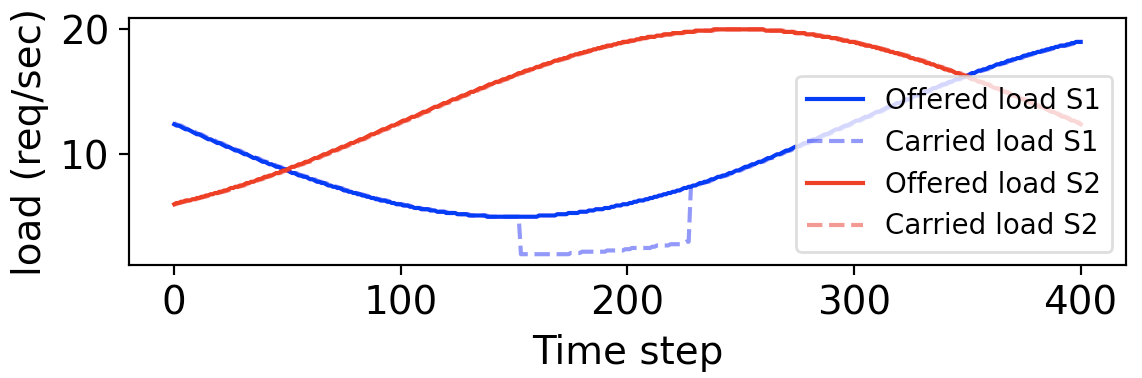}
 \caption{The offered and carried load for Scenario 2, on the simulator for the sinusoidal load pattern.}
 \label{fig:mo3_eval_sin_sim_ld}
\end{figure}

Due to lack of space, we do not include the detailed evaluation of Scenarios 2 and 3, which relate to management objectives 2 and 3. One notable observation is that the RL agent often takes different actions for the same load pattern when the management objective is different. An example can be seen in Figure \ref{fig:mo3_eval_sin_sim_ld}, which shows the offered and carried load under the sinusoidal load pattern on the simulator for management objective 2. This objective prioritizes service $2$ over service $1$. We see that the agent blocks some requests of service $1$ but none of service $2$ between time steps $150$ and $230$. In contrast, Figure \ref{fig:sc1_sim_unseen_ls} shows a case where requests of service $2$ are dropped but none of service $1$, for the same load pattern.  
The reason for the difference is that management objective 1 differentiates between services according to their respective resource consumption, while management objective 2 differentiates according to their respective utility.


Table \ref{tab:anr_results} shows the performance of the RL agent for all three scenarios on the simulator and the testbed. We observe that the agent achieves higher performance on the load pattern it has been trained with (i.e., the random load pattern) than on unseen load (i.e., the sinusoidal load pattern). We expect such a difference in performance because the policy has been optimized for the seen load values. Second, we observe that the performance on the simulator is better than on the testbed for the same load pattern. Similarly, we expect such a difference in performance, because the policy has been computed using the learned system model. On the simulator, the system reacts to an action of the agent according to the system model. On the testbed however, the reaction of the system may differ if the system model is not accurate. Therefore, the difference in performance can be explained by the error estimating the function $f$ in Equation \ref{eq:system_simple}. \\
Overall, our evaluation shows that the agent achieves on the testbed an average $ANR$ value above $0.8$ for unseen load. This performance can be compared to a baseline random policy, where the action parameters $(p_{11}, p_{21}, b_1, b_2)$ are selected uniformly at random from the action space, which results in an $ANR$ value below $0.59$.

We discuss the benefits and drawbacks of including a simulator in our framework. Training the agent on the simulator reduces the policy effectiveness by $0.11$ $ANR$ on average on unseen load (see Table \ref{tab:anr_results}). However, it significantly reduces the overall training time. For the studied scenarios, this requires some $2\times 10^4$ measurement rounds, each lasting a time step of $5$ seconds, resulting in $28$ hours of monitoring time. 

Hyper-parameters tuning and learning effective policy for all scenarios on the testbed needs $2\,499$ hours. However, this time can be reduced to an hour on the simulator. Therefore,
\color{black}
the training time is reduced by a factor of at least $86$. The reduction in time is achieved at the cost of reducing the policy effectiveness by $0.11$ $ANR$. 

\begin{table}[!htbp]
\centering
\caption{Performance of RL agent in all scenarios, running on the simulator and the testbed. Values are given in $ANR$. A value of $1$ means optimal performance where the management objective is met at all times.}
\label{tab:anr_results}
\begin{tabular}{|c|c|c|c|}
\hline
Scenario & Environment & Load pattern & $ANR$\\\hline
1 & simulation & random & 0.99 \\
1 & simulation & sinusoidal & 0.95 \\
1 & testbed & random & 0.85 \\
1 & testbed & sinusoidal & 0.90 \\\hline
2 & simulation & random & 0.96 \\
2 & simulation & sinusoidal & 0.91 \\
2 & testbed & random & 0.81 \\
2 & testbed & sinusoidal & 0.80 \\\hline
3 & simulation & random & 0.98 \\
3 & simulation & sinusoidal & 0.97 \\
3 & testbed & random & 0.85 \\
3 & testbed & sinusoidal & 0.80 \\\hline

\end{tabular}
\end{table} 

\color{black}

\section{Related work}
\label{sec:related_work}

Our work broadly fits into the context of \textit{self-adaptive systems} (see surveys \cite{morandini2008towards}\cite{weyns2012survey}\cite{krupitzer2015survey}), \textit{autonomic computing and networking} (see e.g., \cite{kephart2003vision}\cite{parashar2004autonomic}), and the recent areas of \textit{self-driving systems} (e.g., \cite{schneider2020self}) and \textit{intent-based networking} (see e.g., \cite{rafiq2020intent}\cite{campanella2019intent}). The system design our framework gives rise to is self-adaptive since the agent policy drives the adaptation of the system state, it is autonomous because this adaptation is performed without human interaction, it is self-driving in the sense that it aims to navigate towards a specific goal, and it is intent-based because it is driven by an end-to-end management objective. 

In this paper, we focus on management objectives that relate to the performance of application services on a service mesh. Related to our research are works that make use of RL for resource allocation or works that study resource management in the context of a service mesh.     

In \cite{gari2021reinforcement} Gar{\'\i} et al. present a survey on application-level use cases related to the cloud. The included papers address two issues, namely horizontal and vertical scaling as well as scheduling. The primary approach is model-based or model-free reinforcement learning, and the stated objectives include resource utilization (e.g., CPU, memory, network, and power), SLA violation cost, makespan, waiting time, and response time. Interestingly, all papers assume a VM layer and discuss scaling and scheduling actions related to VMs. 
Seven papers surveyed in \cite{cardellini2019self} apply reinforcement learning to automate resource allocation, scheduling, and container migration. Many of them address the problem of long training times by combining reinforcement learning with other methods like queuing.
\begin{table*}
\centering
\caption{Related work}
\label{tab:related_work}
\scalebox{1}{
\begin{tabular}{|l|c|c|c|c|c|}
\hline
\multirow{2}{*}{Paper}& Control & \multirow{2}{*}{RL System model} &  \multirow{2}{*}{Service abstraction} & Resource management  & Simulation \\
& through RL &  &  & function & Implementation \\\hline
\cite{rossi2020self}, Rossi 2020 & yes & learned from measurements& service mesh & scaling & simulation\\\hline
\cite{chinchali2018cellular}, Chinchal 2018 & yes & learned from measurements& cellular network & routing & simulation \\\hline
\cite{schneider2021self}, Schneider, 2021 & yes & simulation study only & service chain & routing and placement & simulation\\\hline
\cite{garg2021heuristic}, Grag 2021 & yes & simulation study only & service mesh & placement & simulation \\\hline
\cite{ashok2021leveraging}, Ashok 2021 & no & - & service mesh & cross-layer routing & simulation \\\hline
\cite{rajib2022lightweight}, Rajib Hossein 2022 & no & - & service mesh & scaling & implementation\\\hline
\cite{yu2020microscaler}, Yu 2020 & no & - & service mesh & scaling &  simulation \\\hline
\cite{lin2021client}, Lin 2021 & no & - & service mesh & orchestration & implementation  \\\hline

\multirow{2}{*}{This paper} & \multirow{2}{*}{yes} & \multirow{2}{*}{learned from measurements} & \multirow{2}{*}{service mesh} & \multirow{2}{*}{routing and blocking} & simulation \\
&&&&&and implementation\\\hline
\end{tabular}
}
\end{table*} 

Table \ref{tab:related_work} lists recent works that relate to the discussion in this paper. The table categorizes the works based on the applied approach, the service abstraction, the resource management function, and the environment in which the agent learns the policy. Many papers use reinforcement learning to control resource management functions \cite{rossi2020self}\cite{schneider2021self}\cite{chinchali2018cellular}\cite{garg2021heuristic}. Also, while all papers use a directed graph as service abstraction, some of them  specifically consider a microservice-based service mesh, like us in this work \cite{rossi2020self}\cite{schneider2021self}\cite{rajib2022lightweight}\cite{yu2020microscaler}\cite{lin2021client}\cite{garg2021heuristic}\cite{ashok2021leveraging}. Some works are based on implementation \cite{rajib2022lightweight}\cite{lin2021client}, while the rest use only simulation studies for evaluation \cite{rossi2020self}\cite{schneider2021self}\cite{yu2020microscaler}\cite{chinchali2018cellular}\cite{garg2021heuristic}\cite{ashok2021leveraging}. \cite{chinchali2018cellular} is similar to our work in the sense that the system model is learned from measurements. \cite{schneider2021self} is close to our approach since the authors consider several end-to-end performance objectives. 

All works in Table \ref{tab:related_work} study one or more resource management functions, such as scaling, placement, or routing, for a particular use case. When designing a function, the authors associate a performance objective like restricting end-to-end delays for a routing function. We take a different approach with our framework: we start with an end-to-end performance objective and direct the set of resource management functions to jointly meet it. The RL agent takes a combination of actions to meet the objective, whereby one type of action corresponds to one resource management function. For instance, in the scenarios we study in this paper, an end-to-end delay objective is achieved through request routing and blocking. It could as well be achieved through scaling or through a combination of all three functions. In this sense, our approach is top-down and more general, while virtually all related work we are aware of is bottom-up and use-case specific.

\section{Conclusions and future work}
\label{sec:conclusion}
We demonstrated how end-to-end management objectives under varying load can be met through periodic control actions performed by a reinforcement learning agent. By computing near-optimal control policies on a simulator, effective control on a testbed could be achieved in several scenarios with applications on a service mesh, which suggests that the approach and framework we propose is practical. As mentioned in the introduction section, we kept the setup and scenarios simple in order to focus on the totality of the framework. Our planned work includes refinement and extension of the framework towards applicability in practice.
 \\
a) \emph{Practical mesh topologies}: the mesh topology considered in this work is minimal (Figure \ref{fig:service_mesh_sub}). We plan to work with applications that require larger, more complex topologies, which we believe can be handled within our framework. \\
b) \emph{Practical applications:} The two applications used in this work are simple information services with trivial functionality. We will test our approach with more complex, practical applications. \\  
c) \emph{Multistage decision problem:} In this work, the RL agent solves a one-stage decision problem. For a given state, it selects a single control action to bring the system into a state that meets the given management objective. When extending the space of possible control actions to include actions whose settling time is larger than the current control interval (e.g., horizontal scaling), the agent must solve a more complex, multi-stage problem. \\ 
d) \emph{Extending the system state with environment metrics:} In the current work, the system state does not include environment variables like CPU, memory, or network utilization. We made sure that the testbed is well provisioned and no other services or tasks are running during experiments. In practical scenarios, this may not be the case and more complex system models will be required. In our previous research on QoS estimation, we obtained good results for such models on a a similar testbed \cite{stadler2017learning}\cite{yanggratoke2018service}. \\
e) \emph{Increasing sample efficiency for learning the system model and training a policy:} One way to significantly reduce the learning time or, equivalently, the number of samples needed to obtain effective models and policies is through reducing the input space of the system model. We plan to achieve this by exploiting symmetry in the mesh topology and the testbed setup. 
\\
f) \emph{Online policy adaptation:} In our current approach, system model and control policy are learned sequentially and in an offline fashion. In a operational environment, however, the policy must adapt at runtime in response to system failures or changes to the system configuration. We plan to extend our framework towards online learning of the system model with periodic updates. These updates are read by the simulator, which continuously produces updated control policies and sends them to the RL agent on the testbed. 

\section*{Acknowledgements}
The authors are grateful to Andreas Johnsson, Farnaz Moradi, Jalil Taghia, Xiaou Lan, and Hannes Larsson with Ericsson Research for fruitful discussion around this work. The authors thank Kim Hammar and Xiaoxuan Wang at KTH for their helpful comments on a draft of this paper. This research has been partially supported by the Swedish Governmental Agency for Innovation Systems, VINNOVA, through project ANIARA. 
\balance
\bibliographystyle{IEEEtran}
\bibliography{cnsm}

\end{document}